\documentclass{llncs}

\usepackage{color,soul}
\usepackage{gensymb}
\usepackage{amsmath}
\usepackage{algorithm}
\usepackage[noend]{algorithmic}
\usepackage{paralist}
\usepackage{graphicx}
\usepackage{multirow}

\usepackage[caption=false,position=top]{subfig}

\usepackage[normalem]{ulem}

\usepackage{tikz}

\begin{document}

\title{UOLO - automatic object detection and segmentation in biomedical images}
\titlerunning{UOLO - object detection and segmentation}
\author{Teresa Ara\'{u}jo*\inst{1,2} \and Guilherme Aresta*\inst{1,2} \and
Adrian Galdran\inst{1} \and Pedro Costa\inst{1} \and Ana Maria Mendon\c{c}a\inst{1,2} \and Aur\'{e}lio Campilho\inst{1,2} \\ \smallskip \footnotesize{*these authors contributed equally to this work}}
\authorrunning{Teresa Ara\'{u}jo et al.} 
\institute{INESC TEC - Institute for Systems and Computer Engineering, Technology and Science, Porto, Portugal,\\
\email{\{tfaraujo,guilherme.m.aresta,adrian.galdran,pvcosta\}@inesctec.pt},\\ 
\and
Faculdade de Engenharia da Universidade do Porto, Porto, Portugal \\
\email{\{amendon,campilho\}@fe.up.pt}}

\maketitle           

\begin{abstract}
We propose UOLO, a novel framework for the simultaneous detection and segmentation of structures of interest in medical images. 
UOLO consists of an object segmentation module which intermediate abstract representations are processed and used as input for object detection.
The resulting system is optimized simultaneously for detecting a class of objects and segmenting an optionally different class of structures. UOLO is trained on a set of bounding boxes enclosing the objects to detect, as well as pixel-wise segmentation information, when available. 
A new loss function is devised, taking into account whether a reference segmentation is accessible for each training image, in order to suitably backpropagate the error. 
We validate UOLO on the task of simultaneous optic disc (OD) detection, fovea detection, and OD segmentation from retinal images, achieving state-of-the-art performance on public datasets. 
\end{abstract}

\setcounter{footnote}{0}
\section{Introduction}

Detection and segmentation of anatomical structures are central medical image analysis tasks since they allow to delimit Regions-Of-Interest (ROI), create landmarks and improve feature collection. In terms of segmentation, Deep Fully-Convolutional (FC) Neural Networks (NNs) achieve the highest performance on a variety of images and problems. Namely, U-Net \cite{Ronneberger2015} has become a reference model -- its autoencoder structure with skip connections enables the propagation from the encoding to the decoding part of the network, allowing a more robust multi-scale analysis while reducing the need for training data. 

Similarly, Deep Neural Networks (DNNs) have become the technique of choice in many medical imaging detection problems. The standard approach is to use networks pre-trained on large datasets of natural images as feature extractors of a detection module. For instance, Faster-R CNN~\cite{Ren2017} uses these features to identify ROIs via a specialized layer.
ROIs are then pooled, rescaled and supplied to a pair of Fully-Connected NNs responsible for adjusting the size and label the bounding boxes. Alternatively, YOLOv2~\cite{Redmon2016} avoids the use of an auxiliary ROI proposal model by directly using region-wise activations from pre-trained weights to predict coordinates and labels of ROIs.

When a ROI has been identified, the segmentation of an object contained on it becomes much easier. For this reason, the combination of detection and segmentation models into a single method is being explored. For instance, Mask-R CNN~\cite{He2017} extends Faster-R~CNN with the addition of FC layers after its final pooling, enabling a fine segmentation without a significant computational overhead. In this architecture, the segmentation and detection modules are decoupled, \textit{i.e.} the segmentation part is only responsible for predicting a mask, which is then labeled class-wise by the detection module. However, despite the high performance achieved by Mask-R~CNN in computer vision, its application to medical image analysis problems remains limited. This is due to the large requirement of data annotated at a pixel level, which is usually not available in medical applications.

\begin{figure}[t]
\centering
\includegraphics[width=0.65\textwidth]{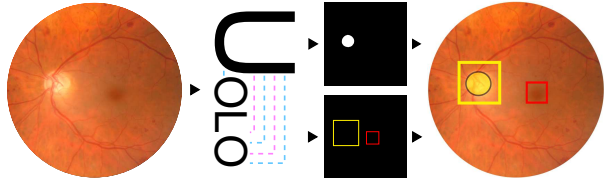}

\caption{Using UOLO for fovea detection and optic disc detection and segmentation.\label{fig:uolo_pipeline}}
\end{figure}

In this paper we propose UOLO (Fig.~\ref{fig:uolo_pipeline}), a novel architecture that performs simultaneous detection and segmentation of structures of interest in biomedical images. 
UOLO harvests the best of its individual detection and segmentation modules to allow robust and efficient predictions even when few training data is available. Moreover, training UOLO is simple since the entire network can be updated during back-propagation.
We experimentally validate UOLO on eye fundus images for the joint task of fovea (FV) detection, optic disc (OD) detection, and OD segmentation, where we achieve state-of-the-art performance. 

\section{UOLO framework}
\label{sec:methods}

\subsection{Object segmentation module}

For object segmentation we consider an adapted version of the U-Net network presented in~\cite{Ronneberger2015}. U-Net is composed of FC layers organized on an auto-encoder scheme, which allows to obtain an output of the same size of the input, thus enabling pixel-wise predictions. Skip connections between the encoding and decoding parts are used for avoiding the information loss inherent to encoding. The model's upsampling path includes a large number of feature channels with the aim of propagating the multi-scale context information to higher resolution layers. Ultimately, the segmentation prediction results from the analysis of abstract representations of the images from multiple scales, with the majority of the relevant classification information being available on the decoder portion of the network due to the skip connections.
We modify the network by adding batch normalization after each convolutional layer, and replacing the pooling layers by convolutions with stride. 
The soft intersection over union (IoU) is used as loss:
\begin{equation}
\mathcal{L}_{\text{U-Net}} = 1 - \text{IoU} = 1 - \frac{\sum{I_t\circ I_p}}{\sum{(I_t+I_p)-\sum{I_t\circ I_p}}},
\label{eq:loss_unet}
\end{equation}
where $I_t$ and $I_p$ are the ground truth mask and the soft prediction mask, respectively, and ${\circ}$ is the Hadamard product.

\par

\subsection{Object detection module}

For object detection we take inspiration from YOLOv2~\cite{Redmon2016}, a network composed of: 
\begin{inparaenum}[1)]
\item{a DNN that extracts features from an image ($F_{\text{YOLO}}$);}
\item{a feature interpretation block that predicts both labels and bounding boxes for the objects of interest ($D_{\text{YOLO}}$).}
\end{inparaenum}
YOLOv2 assumes that every image's patch can contain an object of size similar to one of various template bounding boxes (or \textit{anchors}) computed \textit{a priori} from the objects' shape distribution in the training data.

\par
Let the output of $F_{\text{YOLO}}$ be a tensor $F$ of shape $S\times S\times N$, where $S$ is the dimension of the spatial grid and $N$ is the number of maps. $F_{\text{YOLO}}$ convolves and reshapes $F$ into $Y$, a tensor of shape $S\times S\times A \times (C+5)$, where $A$ is the number of anchors, $C$ is the number of object classes, and 5 is the number of variables to be optimized: center coordinates $x$ and $y$, width $w$, height $h$, and the confidence $c$ (how likely is the bounding box to be an object) of the bounding boxes. For each anchor $A_k$ in $Y$, the value of each feature map element $m_{i,j}$ is responsible for adjusting a property of the predicted bounding box $\hat{b}$,
\begin{align}
(\hat{b}_x,\hat{b}_y) &= (\sigma(\hat{x})+x_{i,j,k},\sigma(\hat{y})+y_{i,j,k})\nonumber\\
(\hat{b}_w,\hat{b}_h) &= (w_{i,j,k}e^{\hat{w}},h_{i,j,k}e^{\hat{h}})\\
\text{confidence} &= \sigma(\hat{c})\nonumber
\end{align}
where $\sigma$ is a sigmoid function. YOLOv2 is trained by optimizing the loss function:
\begin{equation}
\mathcal{L}_{\text{YOLO}} = \lambda_1\mathcal{L}_{\text{centers}}+\lambda_2\mathcal{L}_{\text{dimensions}}+\lambda_3\mathcal{L}_{\text{confidence}}+\lambda_4\mathcal{L}_{\text{classes}}
\label{eq:loss_yolo}
\end{equation}
where $\lambda_i$ are predefined weighting factors, $\mathcal{L}_{\text{centers}}$, $\mathcal{L}_{\text{dimensions}}$ and $\mathcal{L}_{\text{confidence}}$ are mean squared errors, and $\mathcal{L}_{\text{classes}}$ is the cross-entropy loss. Each loss term penalizes a different error:
\begin{inparaenum}[1)]
\item $\mathcal{L}_{\text{centers}}$ penalizes the error in the center position of the cells;
\item $\mathcal{L}_{\text{dimensions}}$ penalizes the incorrect size, \textit{i.e.} height and width, of the bounding box; 
\item $\mathcal{L}_{\text{confidence}}$ penalizes the incorrect prediction of a box presence;
\item $\mathcal{L}_{\text{classes}}$ penalizes the misclassification of the objects.
\end{inparaenum}

\subsection{UOLO for joint object detection and segmentation}

\begin{figure}[b!]
	\includegraphics[width=1\textwidth]{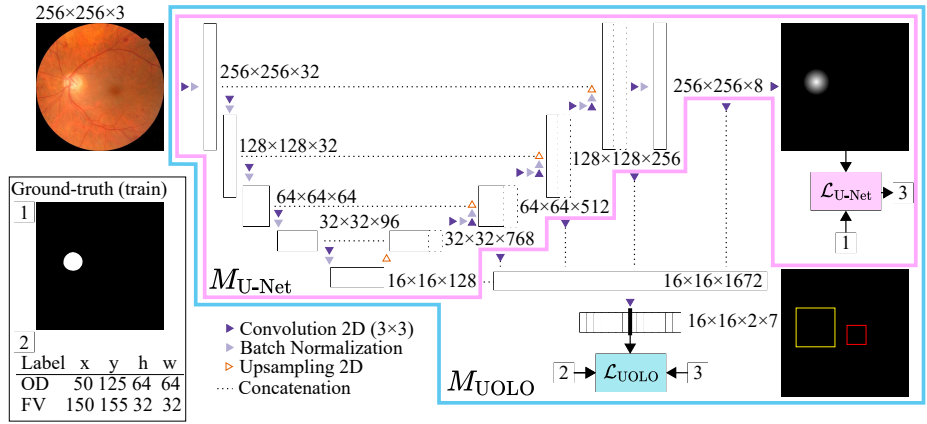}
	\caption{UOLO framework, nesting an U-Net responsible for segmentation and feature extraction for an YOLOv2-based detector.
$M_{\text{U-Net}}$: U-net part; $M_{\text{UOLO}}$: full UOLO.  \label{fig:uolo}}	
\end{figure}

UOLO framework for object detection and segmentation is depicted in Fig.~\ref{fig:uolo}, where the segmentation module itself is used as a feature extraction module, adopting the role of $F_\text{YOLO}$, and serving as input for the localization module $D_\text{YOLO}$. The intuition behind this design is that the abstract representation learned by the decoding part of U-Net contains multi-scale information that can be useful not only to segment objects, but also to detect them. In addition, the class of objects that UOLO can detect is not limited to those for which segmentation ground-truth is available.\par

Let $M_\text{U-Net}$ be an U-Net-like network that, given pairs of images and binary masks, can be trained for performing segmentation by minimizing $\mathcal{L}_{\text{U-Net}}$ (Eq.~\ref{eq:loss_unet}). $M_{\text{U-Net}}$ has a second output corresponding to the concatenation of the downsampled decoding maps with its bottle neck (last encoder layer). The resulting tensor corresponds to a set of multi-scale representations of the original image that are supplied to the object detection block $D_{\text{YOLO}}$, which, by its turn, can be optimized via $\mathcal{L}_{\text{YOLO}}$, defined in Eq.~\ref{eq:loss_yolo}. 
$D_\text{YOLO}$ and $M_\text{U-Net}$ are then merged by concatenation into $M_\text{UOLO}$, a single model that can be optimized by minimizing the addition of the corresponding loss functions:

\begin{equation}
\mathcal{L}_{\text{UOLO}} = \mathcal{L}_\text{YOLO}+\mathcal{L}_\text{U-Net}
\label{eq:loss_uolo}
\end{equation}

\begin{algorithm}[tb]
  \caption{Loss computation scheme of UOLO. \small{$M_{\text{U-Net}}$: U-net part from the UOLO model; $M_{\text{UOLO}}$: full UOLO model; $b_\text{det}$: batches of images with objects' bounding boxes ground truth; $b_\text{seg}$: batches of images with segmentation ground truth.}} 
    \begin{algorithmic}
      \STATE $\mathcal{L}_{\text{U-Net}} \leftarrow 1$
         \FOR {\textbf{each} training step}    
          \STATE $M_{\text{UOLO}} \leftarrow \texttt{train}(M_{\text{UOLO}}, b_\text{det},\mathcal{L}_{\text{UOLO}}) $ \COMMENT{train on $n_{det}$ batches from $b_\text{det}$, backpropagating $\mathcal{L}_\text{UOLO}$}; 
          \STATE $\texttt{update}(\mathcal{L}_{\text{YOLO}}) $

          \smallskip
          \STATE $M_{\text{U-Net}} \leftarrow \texttt{train}(M_{\text{U-Net}}, b_\text{seg},\mathcal{L}_\text{U-Net})$  \COMMENT{train on $n_{seg}$ batches from $b_\text{seg}$, backpropagating $\mathcal{L}_\text{U-Net}$}     

           \STATE $\texttt{update}(\mathcal{L}_{\text{U-Net}}) $

           \STATE $\mathcal{L}_{\text{UOLO}} \leftarrow \mathcal{L}_{\text{YOLO}}+\mathcal{L}_{\text{U-Net}}$ 
      \ENDFOR

    \end{algorithmic}
  \label{alg:train_uolo}
\end{algorithm}

Thanks to the straightforward definition of the loss function in Eq.~(\ref{eq:loss_uolo}), $M_\text{UOLO}$ can be trained with a simple iterative scheme detailed in Algorithm~\ref{alg:train_uolo}. In~essence, $\mathcal{L}_\text{U-Net}$ is updated only when segmentation information is available. However, a global weight update is performed at every step based on the prediction error backpropagation.  
Furthermore, the outlined training scheme allows for a different number of strong (pixel-wise) and weak (bounding boxes) annotations, easing its application to medical images.

\section{Experiments and results}
\label{sec:results}

\subsection{Datasets and experimental details}

We test UOLO on 3 public eye fundus datasets with healthy and pathological images:
\begin{inparaenum}[1)]
\item{Messidor~\cite{Decenciere2014} has 1200 images (1440$\times$960, 2240$\times$1488 and 2304$\times$1536 pixels, 45$^{\circ}$ field-of-view (FOV)), 1136 having ground truth (GT) for OD segmentation and FV centers\footnote{\url{http://www.uhu.es/retinopathy}};}
\item{IDRID\footnote{\url{https://idrid.grand-challenge.org/}, available since January 20, 2018} training set has 413 images (4288$\times$2848 pixels, 50$^{\circ}$ FOV) with OD and FV centers and 54 with OD segmentation;}
\item{DRIVE~\cite{Staal2004} has 40 images (768$\times$584 pixels, 45$^{\circ}$ FOV) with OD segmentation\footnote{\url{https://sites.google.com/a/uw.edu/src/useful-links}}.}
\end{inparaenum}
\par
All images are cropped around the FOV (determined via Otsu's thresholding) and resized to 256$\times$256 pixels.
The side of the square GT bounding boxes is set to $32$ and $64$ for the FV and OD following their relative size in the image. 
For training, $n_\text{det}$ and $n_\text{seg}$ (Alg.~\ref{alg:train_uolo}) are set to 256 and 32, respectively. 
Online data augmentation, a mini-batch size of 8, and the Adam optimizer (learning rate of 1e-4) were used for training, while 25\% of the data was kept for validation.
The bounding box with highest confidence for each class is kept. 
The predicted soft segmentations are binarized using a threshold of 0.5.
\par
The OD segmentation is evaluated with IoU and Sorensen-Dice coefficient overlap metrics.
The detection is evaluated in terms of mean euclidean distance (ED) between the prediction and the GT. We also evaluate ED relatively to the OD radius, $\bar{D}$~\cite{Gegundez-Arias2013,Aquino2014}. Finally, detection success, $S_{1R}$, is assessed using the maximum distance criteria of 1 OD radius.

\subsection{Results and discussion}

\begin{table}
\setlength{\tabcolsep}{3pt}
\renewcommand{\arraystretch}{1.09}
\centering

\caption{UOLO performance on optic disc (OD) detection and segmentation and fovea (FV) detection. $n$: number of training images for detection and segmentation.
\label{tab:exps}}
\begin{tabular}{ll|cc|cc|cc|ccl}
\multicolumn{2}{l|}{\textbf{Datasets}}         & \multicolumn{2}{c|}{\textbf{$\mathbf{n}$}}                                                                                                             & \multicolumn{2}{c|}{\textbf{OD seg.}} & \multicolumn{2}{c|}{\textbf{OD det.}} & \multicolumn{2}{c}{\textbf{FV det.}} &  \\ \cline{1-10}
\textbf{Train}                 & \textbf{Test} & seg. & det.  & IoU     & Dice   & $\bar{D}$       & $S_{1R}$           & $\bar{D}$           & $S_{1R}$                  &  \\ \cline{1-10}
\multicolumn{2}{c|}{\multirow{1}{*}{Messidor}} & 680    & 680   &  0.88   &   0.93   &   0.111        &  99.74   & 0.121 & 99.38\\ 
\multicolumn{2}{c|}{Messidor}                           & 100     &  680    & 0.87   &   0.93    &  0.114                    &  99.74   & 0.114   & 97.89\\
\multicolumn{2}{c|}{IDRID}                       & 30   & 280    &  0.88     &    0.93    &  0.095   &  99.79   &  0.288 &  93.78\\
\multicolumn{1}{l|}{Messidor}  & IDRID         & 852   & 852   &  0.84      &   0.91    &  0.138   &             99.78		&  0.403 & 89.06 &\\
\multicolumn{1}{l|}{Messidor}  & DRIVE       & 852     & 852    &   0.82   &  0.89   &   0.171  &    97.50     &  - & -
\end{tabular}
\end{table}

\begin{figure}[b!]
\centering
\includegraphics[width=\textwidth]{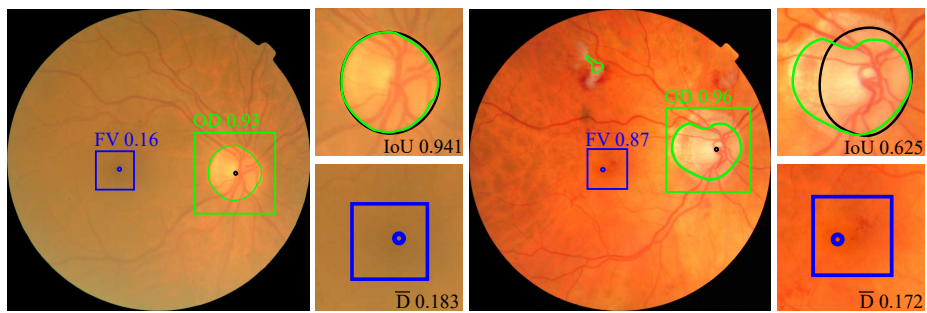} 
\caption{Examples of results of UOLO on Messidor images. Green curve: segmented optic disc (OD), green and blue boxes: predicted OD and FV locations, respectively; black curve: ground truth OD segmentation; black and blue dots: ground truth OD and FV locations, respectively. The object detection confidence is shown next to each box. IoU (intersection over union) and normalized distance ($\bar{D}$) values are also shown.}
\label{fig:results}
\end{figure}

\noindent We evaluate UOLO both inter and intra-dataset-wise. For inter-dataset experiments, UOLO was trained on Messidor and tested in the other datasets whereas for intra-dataset studies stratified 5-fold cross-validation was used. 
We do not extensively optimize the training parameters to verify how robust UOLO is when dealing with segmentation and detection simultaneously. 
Table~\ref{tab:exps} shows the results of UOLO for the OD detection and segmentation and FV detection tasks, Table~\ref{tab:sota} compares our performance with state-of-the-art methods and
Fig.~\ref{fig:results} shows two prediction examples in complex detection and segmentation cases.
\par

UOLO achieves equal or better performance in comparison to the state-of-the-art on both detection and segmentation tasks (IoU $0.88\pm0.09$ on Messidor) in a single step prediction. Furthermore, the proposed network is robust even in inter-dataset scenarios, maintaining both segmentation and detection performances. 
This indicates that the abstract representations learned by UOLO are highly effective for solving the task at hands. It is worth noting that our segmentation and detection performances do not alter significantly even when UOLO is trained with only 15\% of the pixel-wise annotated images. This means that UOLO does not require a significant amount of pixel-wise annotations, easing its application on the medical field, where these are expensive to obtain.
\par
Our results also suggest that UOLO is capable of using multi-scale information (\textit{eg.} relative position to the OD or vessel tree) to perform predictions. For instance, Fig.~\ref{fig:results} shows UOLO's output for two Messidor images, illustrating that the network is capable of detecting the FV in a low contrast scenario. 
On the other hand, the segmentation and detection processes are not completely interdependent, as expected from the proposed training scheme, since the network segments OD confounders outside the detected OD region. Another advantage of UOLO is that these segmentation errors are easily correctable by limiting the pixel-wise predictions to the found OD region. Unlike hand-crafted feature-based methods, UOLO does not require an extensive parameter tunning and it is simple to extend to different applications.

\begin{table}[t]
\setlength{\tabcolsep}{2.3pt}
\renewcommand{\arraystretch}{1.08}
\caption{State-of-the-art for OD detection and segmentation and FV detection. \label{tab:sota}}
\vskip-1em
\subfloat[OD segmentation]{

\begin{tabular}{l|cc|cc}
\multicolumn{1}{c|}{\textbf{Dataset}}         & \multicolumn{2}{c|}{Messidor}  & \multicolumn{2}{c}{DRIVE} \\ \hline
\multicolumn{1}{c|}{\textbf{Method}} & IoU     & Dice   & IoU  & Dice \\ \hline
UOLO                               &  0.88  &  0.93   &   0.82    &    0.89  \\
U-Net                               &  0.88  &  0.93   &   0.81    &    0.88  \\
\cite{Dai2017}              & 0.91   &   -      &  -   &   -   \\
\cite{Dashtbozorg2015}    & 0.89   & 0.94    & -      & -    \\
\cite{Roychowdhury2016}        & 0.84  & - & 0.81  & - \\
\cite{Morales2013}            & 0.82  & -  & 0.72   & - \\
\cite{Salazar-Gonzalez2014}    & -     &  -  &   0.82  & - \\
\end{tabular}
}
\hfill
\subfloat[OD and FV detection]{
\begin{tabular}{l|cc|cc|cc}
\textbf{Task}               & \multicolumn{4}{c|}{\textbf{OD det.}}                                                                                               & \multicolumn{2}{c}{\textbf{FV detection}}  \\ \hline
\textbf{Dataset}            & \multicolumn{2}{c|}{Messidor}     & \multicolumn{2}{c|}{DRIVE}                  & \multicolumn{2}{c}{Messidor}               \\ \hline
\textbf{Method}  & $ED$      & $S_{1R}$    & $ED$      & $S_{1R}$         & $ED$     & $S_{1R}$           \\ \hline
UOLO    			 		&        9.40         &   99.74     &   8.13    &  97.5     & 10.44    &  99.38     \\
YOLOv2   	 &     6.86      &  100    &    7.20    &     97.5      &     9.01 &  100            \\
\cite{Al-Bander2018}		 &      -     & 97        &    -       & -                      &    -   & 96.6                 \\
\cite{Aquino2014}    		&         -   &      -   &           -           &          -             &         16.09             & 98.24                \\
\cite{Gegundez-Arias2013}    &        -       &            -          &           -           &     -     &          20.17        &        98.24         \\
\cite{Aquino2010}    		&      -    &   98.83      &           -           &          -     &    -    & -                \\
\cite{Kamble2017}    		& 23.17     & 99.75        & 15.57       & 100                   & 34.88      & 99.40                \\
\cite{Dashtbozorg2015}   	 &           &  99.75    &    -    &         -           &     - &    -            \\

\end{tabular}

}
\end{table}

We also evaluate U-Net ($\mathcal{M}_\text{U-Net}$, Fig.~\ref{fig:uolo}) for OD segmentation and YOLOv2 (with a pretrained Inceptionv3 as feature extractor) for OD and FV detection (Table~\ref{tab:sota}). The training conditions were set as in UOLO. 
UOLO segmentation performance is practically the same as U-Net, whereas the detection  drops slightly when comparing with YOLOv2, mainly for OD detection. 
However, one has to consider the trade-off between computational burden and performance, since UOLO network has $23~347~063$ parameters, whereas U-Net has $15~063~985$ and YOLOv2 has $21~831~470$, being that for training U-Net and YOLO a total of $36~895~455$ parameters have to be optimized (60\% increase).

\section{Conclusions}

We presented UOLO, a novel network that performs joint detection and segmentation of objects of interest in medical images by using the abstract representations learned by U-Net. Furthermore, UOLO can detect objects from a different class for which segmentation ground-truth is available. 
\par
We tested UOLO for simultaneous fovea detection and optic disk detection and segmentation, achieving state-of-the-art results. 
This network can be trained with relatively few images with segmentation ground-truth and still maintain a high performance. UOLO is also robust to inter-dataset settings, thus showing great potential for applications in the medical image analysis field.

\subsubsection*{Acknowledgements}

T.~Ara\'{u}jo is funded by the FCT grant SFRH/BD/122365/ 2016. G.~Aresta is funded by the FCT grant SFRH/BD/120435/ 2016. 
This work is funded by the ERDF European Regional Development Fund, Operational Programme for Competitiveness and Internationalisation - COMPETE 2020, and by National Funds through the FCT - project CMUP-ERI/TIC/0028/2014.

\bibliographystyle{splncs}
\bibliography{Mendeley.bib}

\end{document}